\RequirePackage{amsmath}
\documentclass[runningheads]{llncs}
\usepackage[T1]{fontenc}
\usepackage{graphicx}
\usepackage{multirow}
\usepackage{capt-of}
\usepackage{amsfonts}
\usepackage{amsmath}
\usepackage{booktabs}
\usepackage{tabularx}
\usepackage{pifont}
\usepackage{xspace}
\usepackage{bm}
\usepackage{paralist}

\usepackage{fancyvrb}
\usepackage{color}
\definecolor{cb_orange}{rgb}{1.0,0.51,0.0}
\definecolor{cb_blue}{rgb}{0.22,0.49,0.72}
\definecolor{cb_green}{rgb}{0.3,0.67,0.29}
\definecolor{cb_red}{rgb}{0.89,0.1,0.11}
\definecolor{cb_purple}{rgb}{0.6, 0.31, 0.64}
\definecolor{cadetgrey}{rgb}{0.57, 0.64, 0.69}

\newcommand{\delete}[1]{}

\newcommand{\prior}[0]{PRIME\xspace}
\newcommand{\ours}[0]{PRIME\texttt{+}\xspace}

\begin{document}
\title{Enhancing Breast Cancer Risk Prediction by Incorporating Prior Images}
\titlerunning{Enhanced Risk Prediction With Prior Images}
%

\author{Hyeonsoo Lee\inst{1} \and
Junha Kim\inst{1} \and
Eunkyung Park\inst{1} \and
Minjeong Kim\inst{1} \and
Taesoo Kim\inst{1} \and
Thijs Kooi\inst{1}
}

\authorrunning{H. Lee et al.}
%
\institute{Lunit Inc, Seoul, Republic of Korea \\
\email{\{hslee, junha.kim, ekpark, mjkim0918, taesoo.kim, tkooi@lunit.io\}}}
\maketitle              
\begin{abstract}

Recently, deep learning models have shown the potential to predict breast cancer risk and enable targeted screening strategies, but current models do not consider the change in the breast over time. In this paper, we present a new method, \ours, for breast cancer risk prediction that leverages prior mammograms using a transformer decoder, outperforming a state-of-the-art risk prediction method that only uses mammograms from a single time point. We validate our approach on a dataset with 16,113 exams and further demonstrate that it effectively captures patterns of changes from prior mammograms, such as changes in breast density, resulting in improved short-term and long-term breast cancer risk prediction. Experimental results show
that our model achieves a statistically significant improvement in performance over the state-of-the-art based model, with a C-index increase from 0.68 to 0.73 (p < 0.05) on held-out test sets.

\end{abstract}

\section{Introduction}
\label{sec:introduction}





Breast cancer impacts women globally~\cite{BreastCancerStatistics} and mammographic screening for women over a certain age has been shown to reduce mortality~\cite{hakama2008cancer,paci2012summary,duffy2020effect}.
However, studies suggest that mammography alone has limited sensitivity~\cite{health2007screening}. To mitigate this, supplemental screening like MRI or 
a tailored screening interval have been explored to add to the screening protocol~\cite{bakker2019supplemental,hussein2023supplemental}. However, these imaging techniques are expensive and add additional burdens for the patient. Recently, several studies~\cite{yala2021toward,yala2022multi,eriksson2022risk} revealed the potential of artificial intelligence (AI) to develop a better risk assessment model to identify women who may benefit from supplemental screening or a personalized screening interval and these may lead to improved screening outcomes.

In clinical practice, breast density and traditional statistical methods for predicting breast cancer risks such as the Gail~\cite{BCRiskTool2011} and the Tyrer-Cuzick models~\cite{tyrer2004breast} have been used to estimate an individual's risk of developing breast cancer. However, these models do not perform well enough to be utilized in practical screening settings~\cite{brentnall2020risk} and require the collection of data that is not always available. Recently, deep neural network based models that predict a patient's risk score directly from mammograms have shown promising results~\cite{brentnall2020risk,eriksson2022risk,yala2021toward,liu2020decoupling,gastounioti2022artificial}. These models do not require additional patient information and have been shown to outperform traditional statistical models. 


When prior mammograms are available, radiologists compare prior exams to the current mammogram to aid in the detection of breast cancer. Several studies have shown that utilizing past mammograms can improve the classification performance of radiologists in the classification of benign and malignant masses \cite{hayward2016improving,varela2005use,roelofs2007importance,sumkin2003optimal}, especially for the detection of subtle abnormalities \cite{roelofs2007importance}. More recently, deep learning models trained on both prior and current mammograms have shown improved performance in breast cancer classification tasks~\cite{park2019screening}. Integrating prior mammograms into deep learning models for breast cancer risk prediction can provide a more comprehensive evaluation of a patient's breast health.


In this paper we introduce a deep neural network that makes use of prior mammograms, to assess a patient's risk of developing breast cancer, dubbed \ours (PRIor Mammogram Enabled risk prediction). We hypothesize that mammographic parechnymal pattern changes between current and prior allow the model to better assess a patient's risk. Our method is based on a transformer model that uses attention~\cite{vaswani2017attention}, similar to how radiologists would compare current and prior mammograms. 

The method is trained and evaluated on a large and diverse dataset of over 9,000 patients and shown to outperform a model based on state-of-the art risk prediction techniques for mammography \cite{yala2021toward}. To the best of our knowledge, this is the first breast cancer risk prediction model which effectively leverages the information from both prior and current mammograms.

\section{Method}
\label{sec:methods}
\subsection{Risk prediction}
\begin{figure*}[t]
\begin{center}
\label{fig:overview}
\includegraphics[width=\textwidth]{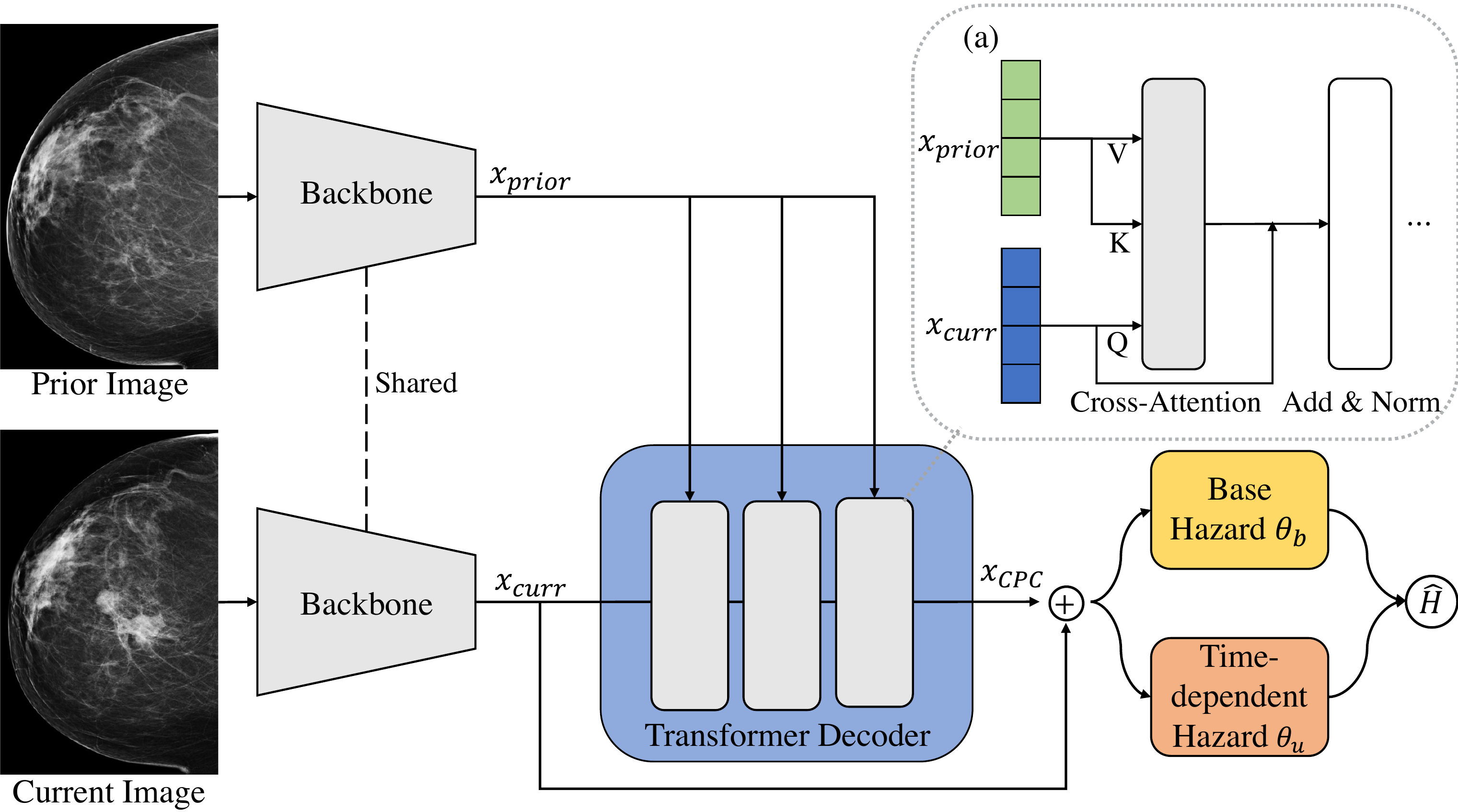}
\caption{We present an improved method for breast cancer risk prediction (\ours) by leveraging prior mammograms. 
A common backbone network extracts features from the prior and current images, resulting in $x_{prior}$ and $x_{curr}$. We find that the transformer decoder effectively fuses relevant information from $x_{prior}$ and $x_{curr}$ to produce $x_{CPC}$.
The base hazard $\theta_{b}$ and time-dependent hazard prediction heads $\theta_{u}$ use the concatenated feature to predict the cumulative hazard function $\hat{H}$. (a) illustrates the interaction between $x_{prior}$ and $x_{curr}$ in the cross-attention module of the transformer decoder.}

\label{fig:method}
\end{center}
\end{figure*}

Survival analysis is done to predict whether events will occur sometime in the future. The data comprises three main elements: features $x$, time of the event $t$, and the occurrence of the event $e$~\cite{katzman2018deepsurv}. For medical applications, $x$ typically represents patient information like age, family history, genetic makeup, and diagnostic test results (e.g., a mammogram). If the event has not yet occurred by the end of the study or observation period, the data is referred to as right-censored.

We typically want to estimate the hazard function $h(t)$, which measures the rate at which patients experience the event of interest at time $t$, given that they have survived up to that point. The hazard function can be expressed formally as the limit of the conditional probability of an event occurring within a small time interval $[t, t+\Delta t)$, given that the time $t$ and has not yet experienced the event:

\begin{equation}
h(t) = \lim_{\Delta t \to 0} \frac{P(T \in [t, t+\Delta t) \mid T \geq t)}{\Delta t}
\end{equation}
The cumulative hazard function $H(t)$ is another commonly used function in survival analysis, which gives the accumulated probability of experiencing the event of interest up to time $t$. This function is obtained by integrating the hazard function over time from 0 to $t$: $ H(t) = \int_0^t h(s) ds $

\subsection{Architecture Overview}
We build on the current state-of-the art MIRAI \cite{yala2021toward} architecture, which is trained to predict the cumulative hazard function.

We use an ImageNet pretrained ResNet-34 \cite{he2016deep} as the image feature backbone. The backbone network extracts features from the mammograms, and the fully connected layer produces the final feature vector $x$. We make use of two additional fully connected layers to calculate base hazard $\theta_{b}$ and time-dependent hazard $\theta_{u}$, respectively.

The predicted cumulative hazard is obtained by adding the base hazard and time-dependent hazard, according to:
\begin{equation}
    \hat{H}(t|x) = \theta_{b}(x) + \sum_{\tau=1}^t \theta_{u_\tau}(x)
\end{equation}

When dealing with right-censored data, we use an indicator function $\delta_i(t)$ to determine whether the information for sample $i$ at time $t$ should be included in the loss calculation or not. This helps us exclude unknown periods and only use the available information. It is defined as follows:

\begin{equation}
    \delta_i(t) =
    \begin{cases}
        1, & \text{if the event of interest occurs for sample $i$ ($e_i=1$)} \\
        1, & \text{if sample $i$ is right-censored at time $t$ ($e_i=0$ and $t<C_i$)} \\
        0, & \text{otherwise}
    \end{cases}
\end{equation}

Here, $e_i$ is a binary variable indicating whether the event of interest occurs for sample $i$ (i.e., $e_i=1$) or not (i.e., $e_i=0$), and $C_i$ is the censoring time for sample $i$, which is the last known time when the sample was cancer-free.

We define the ground-truth $H$ is a binary vector of length $T$, where $T$ is the maximum observation period. Specifically, $H(t)$ is 1 if the patient is diagnosed with cancer within $t$ years and 0 otherwise. We use binary cross entropy to calculate the loss at time $t$ for sample $i$: $\ell_i(t) = -H_i(t) \log \hat{H}_{i}(t)-(1-H_i(t)) \log (1-\hat{H}_{i}(t))$.
The total loss is defined as:

\begin{equation}
    L=\sum_{i=1}^N \sum_{t=1}^T \delta_i(t) \ell_i(t)
\end{equation}


Here, $N$ is the number of exams in the training set. The goal of training the model is to minimize this loss function, which encourages the model to make accurate predictions of the risk of developing breast cancer over time.

\subsection{Incorporating Prior Mammograms}
To improve the performance of the breast cancer risk prediction model, we incorporate information from prior mammograms taken with the same view, using a transformer decoder structure~\cite{vaswani2017attention}. This structure allows the current and prior mammogram features to interact with each other, similar to how radiologists check for changes between current and prior mammograms.

During training, we randomly select one prior mammogram, regardless of when they were taken. This allows the model to generalize to varying time intervals. To pair each current mammogram during inference with the most relevant prior mammogram, we first select the prior mammogram taken at the time closest to the current time. This approach is based on research showing that radiologists often use the closest prior mammogram to aid in the detection of breast cancer~\cite{sumkin2003optimal}.

Next, a shared backbone network is used to output the current feature $x_{curr}$ and the prior feature $x_{prior}$. These features are then flattened and fed as input to the transformer decoder, where multi-head attention is used to find information related to the current feature in the prior feature. The resulting output is concatenated and passed through a linear layer to produce the current-prior comparison feature $x_{CPC}$. The current-prior comparison feature and current feature are concatenated to produce the final feature $x^* = x_{CPC} \oplus x_{curr}$, which is then used by the base hazard network and time-dependent hazard network to predict the cumulative hazard function $\hat{H}$.

\section{Experiments}
\label{sec:experiments}
\subsection{Dataset}
We compiled an in-house mammography dataset comprising 16,113 exams (64,452 images) from 9,113 patients across institutions from the United States, gathered between 2010 and 2021. Each mammogram includes at least one prior mammogram. The dataset has 3,625 biopsy-proven cancer exams, 5,394 biopsy-proven benign exams, and 7,094 normal exams. Mammograms were captured using Hologic (72.3\%) and Siemens (27.7\%) devices. We partitioned the dataset by patient to create training, validation, and test sets. The validation set contains 800 exams (198 cancer, 210 benign, 392 normal) from 400 patients, and the test set contains 1,200 exams (302 cancer, 290 benign, 608 normal) from 600 patients.
All data was de-identified according to the the U.S HHS Safe Harbor Method. Therefore, the data has no  PHI (Protected Health Information) and IRB (Institutional Review Board) approval is not required.

\subsection{Evaluation}

We make use of Uno's C-index~\cite{uno2011c} and the time-dependent AUC~\cite{kamarudin2017time}. The C-index measures the performance of a model by evaluating how well it correctly predicts the relative order of survival times for pairs of individuals in the dataset. The C-index ranges from 0 to 1, with a value of 0.5 indicating random predictions and a value of 1 indicating that the model is perfect. Time-dependent ROC analysis generates an ROC curve and the area under the curve (AUC) for each specific time point in the follow-up period, enabling evaluation of the model's performance over time.
To compare the C-index of two models, we employ the compareC~\cite{kang2015comparing} test, and make use of the DeLong test \cite{delong1988comparing} to compare the time-dependent AUC values. Confidence bounds are generated using bootstrapping with 1,000 bootstraps. 

We evaluate the effectiveness of \ours  by comparing it with two other models: (1) baseline based on MIRAI, a state-of-the art risk prediction method from~\cite{yala2021toward}, and (2) \prior, a model that uses prior images by simply summing $x_{curr}$ and $x_{prior}$ without the use of the transformer decoder. 

\subsection{Implementation Details}

Our model is implemented in Pytorch and trained on four V100 GPUs. We trained the model using stochastic gradient descent (SGD) for 20K iterations with a learning rate of 0.005, weight decay of 0.0001, and momentum of 0.9. We use a cosine annealing learning rate scheduling strategy~\cite{loshchilov2016sgdr}.

We resize the images to 960 $\times$ 640 pixels and use a batch size of 96. To augment the training data, we apply geometric transformations such as vertical flipping, rotation and photometric transformations such as brightness/contrast adjustment, Gaussian noise, sharpen, CLAHE, and solarize. Empirically, we find that strong photometric augmentations improved the risk prediction model's performance, while strong geometric transformations had a negative impact. This is consistent with prior work \cite{liu2020decoupling} showing that risk prediction models focus on overall parenchymal pattern.

\subsection{Results}

\textbf{Ablation Study.}
\begin{table}[t]
\centering
\caption{Ablation analysis on the effectiveness of prior information and transformer decoder. Additional result in bottom row aims to predict unseen risks beyond visible cancer patterns by excluding early diagnosed cancer cases. The ± refers to the 95\% confidence bound.}
\label{table:ablation}
\begin{tabularx}{\linewidth}{|>{\centering\arraybackslash}X*{7}{|>{\centering\arraybackslash}X}|}
\hline
\multicolumn{7}{|c|}{\textit{All Cases}} \\
\hline
\multirow{2}{*}{\textbf{Prior}} & \multirow{2}{*}{\textbf{Decoder}} & \multirow{2}{*}{\textbf{C-index}} & \multicolumn{4}{c|}{\textbf{Time-dependent AUC}} \\
\cline{4-7}
& & & \textbf{1-year} & \textbf{2-year} & \textbf{3-year} & \textbf{4-year} \\
\hline
\ding{55} & \ding{55} & 0.68$_{\pm0.03}$ & 0.70$_{\pm0.05}$ & 0.71$_{\pm0.04}$ & 0.70$_{\pm0.04}$ & 0.71$_{\pm0.09}$ \\
\ding{51} & \ding{55} & 0.70$_{\pm0.03}$ & 0.72$_{\pm0.05}$ & 0.73$_{\pm0.05}$ & 0.74$_{\pm0.04}$ & 0.75$_{\pm0.07}$ \\
\ding{51} & \ding{51} & 0.73$_{\pm0.03}$ & 0.75$_{\pm0.05}$ & 0.75$_{\pm0.04}$ & 0.77$_{\pm0.04}$ & 0.76$_{\pm0.08}$ \\
\hline
\multicolumn{7}{|c|}{\textit{Excluding Cancer Cases with Event Time $<$ 180 Days}} \\
\hline
\multirow{2}{*}{\textbf{Prior}} & \multirow{2}{*}{\textbf{Decoder}} & \multirow{2}{*}{\textbf{C-index}} & \multicolumn{4}{c|}{\textbf{Time-dependent AUC}} \\
\cline{4-7}
& & & \textbf{1-year} & \textbf{2-year} & \textbf{3-year} & \textbf{4-year} \\
\hline
\ding{55} & \ding{55} & 0.63$_{\pm0.04}$ & 0.64$_{\pm0.10}$ & 0.66$_{\pm0.08}$ & 0.64$_{\pm0.06}$ & 0.64$_{\pm0.11}$ \\
\ding{51} & \ding{55} & 0.68$_{\pm0.05}$ & 0.64$_{\pm0.14}$ & 0.73$_{\pm0.08}$ & 0.70$_{\pm0.05}$ & 0.71$_{\pm0.09}$ \\
\ding{51} & \ding{51} & 0.70$_{\pm0.04}$ & 0.68$_{\pm0.13}$ & 0.76$_{\pm0.07}$ & 0.73$_{\pm0.05}$ & 0.71$_{\pm0.10}$ \\
\hline
\end{tabularx}
\end{table}
To better understand the merit of the transformer decoder, we first performed an ablation study on the architecture. Our findings, summarized in Table~\ref{table:ablation}, include two sets of results: one for all exams in the test set and the other by excluding cancer exams within 180 days of cancer diagnosis which are likely to have visible symptoms of cancer, by following a previous study~\cite{yala2021toward}. This latter set of results is particularly relevant as risk prediction aims to predict unseen risks beyond visible cancer patterns. We also compare our method to two other models, the state-of-the-art baseline and \prior models.

As shown in top rows in Table~\ref{table:ablation}, the baseline obtained a C-index of 0.68 (0.65 to 0.71). By using the transformer decoder to jointly model prior images, we observed improved C-index from 0.70 (0.67 to 0.73) to 0.73 (0.70 to 0.76).
The C-index as well as all AUC differences between the baseline and the \ours are all statistically significant (p < 0.05) except the 4-year AUC where we had limited number of test cases.

We observe similar performance improvements when evaluating using cases with at least 180 days to cancer diagnosis.
Interestingly, the C-index as well as time-dependent AUCs of all three methods decreased compared to when evaluating using all cases. 
The intuition behind this result is that mammograms taken near the cancer diagnosis ( < 180 days) likely contain visible signs of cancer and thus the task of risk prediction is easier.
The model must learn patterns of risk, not visible signs of cancer, in order to perform well under this evaluation setting. Our results support this intuition as the performance improvements over the baseline are much more pronounced for longer term risk (3, 4-year AUC) than short term risk (1 year). 
The \prior and \ours models, which incorporate prior mammograms, show high performance for long-term risk prediction (3, 4-year AUC), indicating that considering changes in breast over time contain useful information for breast cancer risk prediction.


Lastly, we empirically confirm that a transformer decoder effectively models spatial relations between prior and current mammograms by demonstrating consistent performance improvements of \ours across both short-term and long-term risk prediction settings.
Our results suggest that incorporating changes in patients using prior mammograms and a transformer decoder improves the performance of breast cancer risk prediction models.

\begin{table}[t]
\centering
\caption{To better understand why the addition of prior images works, we split our test set into two groups based on the mammographic density: change and no change. The first and second row corresponds to performance of the baseline and RP+ model, respectively. Empty cell indicates an insufficient number of cases available for evaluation.}
\label{table:density_change}
\begin{tabularx}{\linewidth}{|>{\centering\arraybackslash}X*{6}{|>{\centering\arraybackslash}X}|}
\hline
\multirow{2}{*}{\textbf{Density chg}} & \multirow{2}{*}{\textbf{C-index}} & \multicolumn{4}{c|}{\textbf{Time-dependent AUC}} \\
\cline{3-6}
& & \textbf{1-year} & \textbf{2-year} & \textbf{3-year} & \textbf{4-year} \\
\hline
\multirow{2}{*}{\textbf{Change}} 
& 0.63$_{\pm0.14}$ & 0.74$_{\pm0.17}$ & 0.66$_{\pm0.18}$ & 0.56$_{\pm0.16}$ & - \\
& 0.75$_{\pm0.10}$ & 0.82$_{\pm0.13}$ & 0.76$_{\pm0.14}$ & 0.74$_{\pm0.14}$ & - \\
\hline
\multirow{2}{*}{\textbf{No change}} 
& 0.69$_{\pm0.03}$ & 0.70$_{\pm0.05}$ & 0.72$_{\pm0.05}$ & 0.72$_{\pm0.04}$ & 0.71$_{\pm0.09}$ \\
& 0.73$_{\pm0.03}$ & 0.74$_{\pm0.05}$ & 0.75$_{\pm0.05}$ & 0.77$_{\pm0.04}$ & 0.76$_{\pm0.08}$ \\
\hline
\end{tabularx}
\end{table}

\textbf{Analysis based on density.}
To better understand why adding prior images improves performance, we divided our test set into subgroups to examine the performance of the baseline model and the \ours model on each of these groups.
Mammographic breast density is one of the most important risk factor to predict breast cancer~\cite{veronesi2005goldhirsch,lee2017risk}.
Women with dense breasts have a four-to six-fold higher risk of breast cancer~\cite{boyd2013mammographic}.
The addition of mammographic breast density has improved the performance traditional breast cancer risk models~\cite{brentnall2015mammographic} and can therefore help us understand why the addition of prior images works. 

Mammographic breast density was determined using the Breast Imaging Reporting and Data System (BI-RADS) composition classification. BI-RADS category A,B are defined as fatty breasts and BI-RADS category C, D are classified as dense breasts.
To determine the density category, we employed an internally developed density prediction model, as most exams lack BI-RADS ground truth. This model achieved an accuracy of 0.81 on the internal density validation set.

We categorized the exams into two groups based on changes in density: "change" and "no change".
Density change was defined according to whether the BI-RADS category is changed in the current image as compared to the prior image.
As shown in Table~\ref{table:density_change}, the baseline model performs poorly for "change", with a C-index of 0.63 (0.49 to 0.77), especially for long-term risk prediction, with 3-year AUC of 0.56 (0.40 to 0.72). This suggests that the baseline model have limitations in accurately predict long-term risk when there is a density change from the prior exam.
However, \ours is able to predict long-term risk accurately even when a density change has occurred (3-year AUC = 0.74 (0.60 to 0.88)), by learning to refer previous exams properly. This demonstrates the potential usefulness of incorporating past mammogram information into breast cancer risk prediction models.
Thus, we believe that incorporating prior exams is important to identify changes in texture which are important for long term risk prediction.


\begin{table}[t]
\centering
\caption{In order to assess the performance of the models on varying levels of breast density, a critical risk factor, we divided our test set into two groups based on mammographic density: fatty and dense.}
\label{table:density}
\begin{tabularx}{\linewidth}{|>{\centering\arraybackslash}X*{6}{|>{\centering\arraybackslash}X}|}
\hline
\multirow{2}{*}{\textbf{Density}} & \multirow{2}{*}{\textbf{C-index}} & \multicolumn{4}{c|}{\textbf{Time-dependent AUC}} \\
\cline{3-6}
& & \textbf{1-year} & \textbf{2-year} & \textbf{3-year} & \textbf{4-year} \\
\hline
\multirow{2}{*}{\textbf{Fatty}} 
& 0.70$_{\pm0.04}$ & 0.73$_{\pm0.06}$ & 0.74$_{\pm0.05}$ & 0.70$_{\pm0.05}$ & 0.70$_{\pm0.10}$ \\
& 0.74$_{\pm0.04}$ & 0.76$_{\pm0.06}$ & 0.76$_{\pm0.05}$ & 0.78$_{\pm0.05}$ & 0.76$_{\pm0.08}$ \\
\hline
\multirow{2}{*}{\textbf{Dense}} 
& 0.68$_{\pm0.06}$ & 0.66$_{\pm0.09}$ & 0.68$_{\pm0.09}$ & 0.71$_{\pm0.08}$ & 0.65$_{\pm0.21}$ \\
& 0.71$_{\pm0.05}$ & 0.72$_{\pm0.08}$ & 0.73$_{\pm0.08}$ & 0.72$_{\pm0.08}$ & 0.72$_{\pm0.25}$ \\
\hline
\end{tabularx}
\end{table}
Lastly, we divided the exams based on the level of breast density, with a fatty group consisting of density A and B, and a dense group consisting of density C and D. Both the baseline and \ours performs better in fatty group than dense group. We suspect this is because deep neural networks generally work better on low density images given that visual cues of cancer in images with lower breast density are more clearly visible.

\section{Conclusion}
\label{sec:conclusion}
In this paper, we introduce a novel breast cancer risk prediction method, \ours, which incorporates prior mammograms with a transformer decoder to capture changes in breast tissue over time. By doing so, we achieve high performance for both short-term and long-term risk prediction. Our extensive experiments on a dataset of 16,113 exams show that \ours outperformed a model based on the state-of-the art for breast cancer risk prediction \cite{yala2021toward}. The method performed particularly well in cases where there was a change in breast density from the previous exam. We believe that our method has the potential to improve breast cancer risk prediction models and ultimately contribute to earlier detection of the disease.

%
%
%
\bibliographystyle{splncs04}
\bibliography{refs}

\end{document}